\documentclass[sigconf]{acmart}
\AtBeginDocument{%
  }

\setcopyright{acmlicensed}
\copyrightyear{2026}
\acmYear{2026}
\acmDOI{XXXXXXX.XXXXXXX}
\acmConference[ACM AI Leadership Summit '26]{Visionary Track}{August 30--September 02, 2026}{Atlanta, GA, USA}
\acmISBN{978-1-4503-XXXX-X/2018/06}

\usepackage{pifont}

\newcommand{\rqnum}[1]{\ding{\the\numexpr181+#1\relax}}

\begin{document}

\newcommand{\sk}[1]{{\color{purple}{\{subangkar: #1\}}}}
\newcommand{\Alireza}[1]{{\authnote{\textcolor{blue}{Alireza}}{\textcolor{blue}{#1}}}}
\newcommand{\added}[1]{{#1}}

\title{Securing Agentic AI: From Per-Action Checks to Trajectory Assurance}
\subtitle{\textit{This work has been accepted to the ACM AI Leadership Summit 2026, Visionary Track}}


\author{Alireza Lotfi}
\authornote{Both authors contributed equally to this research.}
% \orcid{0009-0000-5694-0278}
\affiliation{%
  \institution{Purdue University}
  \department{Department of Computer Science}
  \streetaddress{305 North University Street}
  \city{West Lafayette}
  \state{IN}
  \country{USA}}
\email{lotfia@purdue.edu}

\author{Subangkar Karmaker Shanto}
\authornotemark[1]
% \orcid{0009-0000-5694-0278}
\affiliation{%
  \institution{Purdue University}
  \department{Department of Computer Science}
  % \streetaddress{305 North University Street}
  \city{West Lafayette}
  \state{IN}
  \country{USA}}
\email{sshanto@purdue.edu}

\author{Imtiaz Karim}
% \orcid{0000-0002-4029-7051}
\affiliation{%
  \institution{University of Texas at Dallas}
  \department{Department of Computer Science}
  % \streetaddress{305 North University Street}
  \city{Richardson}
  \state{TX}
  \country{USA}}
\email{imtiaz.karim@utdallas.edu}

\author{Elisa Bertino}
% \orcid{0000-0002-4029-7051}
\affiliation{%
  \institution{Purdue University}
  \department{Department of Computer Science}
  % \streetaddress{305 North University Street}
  \city{West Lafayette}
  \state{IN}
  \country{USA}}
\email{bertino@purdue.edu}

% \renewcommand{\shortauthors}{Lotfi and shanto et al.}
\renewcommand{\shortauthors}{Lotfi, Shanto, et al.}

\begin{abstract}
Autonomous agents are increasingly used to execute consequential tasks in environments governed by operational constraints, organizational policies, regulatory requirements, and technical standards. Their safety is therefore determined not by the correctness of individual actions, but by whether their overall behavior remains consistent with the rules and invariants of the systems in which they operate. As large language model (LLM)-based agents become more autonomous and increasingly delegate tasks across organizational boundaries, securing them evolves from a single challenge into a broad and interconnected landscape spanning the entire agentic stack. At the single-agent level, untrusted inputs through prompts, memory, retrieved knowledge, and tool interfaces create attack surfaces. In multi-agent settings, delegation and communication introduce challenges related to identity, trust, capability control, and decision transparency, while the underlying model routing and execution control plane remains vulnerable to manipulation and to unverified model provenance. Perhaps the most fundamental challenge is behavioral containment: sequences of individually permissible actions may collectively violate system-level constraints and safety invariants. At the broader level, supply-chain integrity, provenance, accountability, and end-to-end observability remain largely open problems. A common principle unifies these directions: security must become a verifiable property of the architectures, protocols, and runtimes that govern agent behavior, rather than an optional layer of guidance. Charting these challenges provides a roadmap toward trustworthy autonomous agent deployment.
\end{abstract}

% \received[accepted]{31 July 2026}

\maketitle

% \begin{center}
% \begin{minipage}{0.95\textwidth}
% \centering
% \small\textit{This work has been accepted to the ACM AI Leadership Summit 2026, Visionary Track.}
% \end{minipage}
% \end{center}

\section{Introduction}
\label{sec:intro}

\begin{figure}
    \centering
    \includegraphics[trim=0 6 0 0, clip, width=0.90\linewidth]{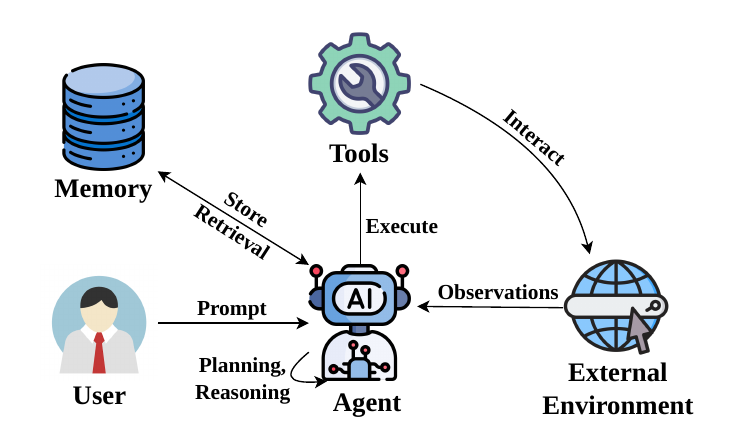}
    \caption{Agentic AI architecture. An agent plans, uses memory and tools, and interacts with users and environments.}
    \label{fig:architecture}
    % \vspace{-.60cm}
\end{figure}

Autonomous agents powered by large language models (LLMs) are increasingly entrusted with critical tasks across enterprise and cyber-physical environments, including healthcare, finance, telecommunications, and critical infrastructure. Unlike traditional AI systems that primarily provide recommendations, these agents plan, reason, invoke tools, interact with external systems, and increasingly collaborate with other agents to accomplish complex objectives. As a result, a single agent may issue thousands of tool calls while a human operator reviews only a handful of decisions. In many of these deployments, agent behavior is constrained by organizational policies, regulatory requirements, or technical standards that specify not only which individual actions are permitted, but also the behavioral envelope within which an entire sequence of actions must remain. Figure~\ref{fig:architecture} illustrates the core components and interaction flows of such an agent. Also agents rarely operate in isolation. Through emerging interoperability protocols they discover, coordinate with, and delegate tasks to other agents, allowing a single workflow to span multiple autonomous entities across organizational and vendor boundaries. This makes securing the agentic systems a broad and interconnected research landscape.
Therefore, securing agentic AI has attracted growing attention from academia and industry~\cite{survey_bertino}. Recent efforts have begun to structure this landscape: the OWASP Agentic Security Initiative identifies fifteen categories of agentic threats~\cite{owasp_asi}, the Cloud Security Alliance's MAESTRO framework organizes the agentic stack into seven layers~\cite{maestro}, and Microsoft's updated failure-mode taxonomy highlights emerging risks ranging from supply-chain compromise to goal hijacking~\cite{ms_agentic_taxonomy}. Rather than introducing another taxonomy, our focus is to provide a \emph{lens} for analyzing the agentic security landscape through the demands of governed deployment, where the intended behavior is an explicit written artifact rather than a matter of designer judgment. We organize the major open problems into eleven research directions spanning the entire agentic stack from a single agent's reasoning, memory, and tools to multi-agent collaboration, model routing, behavioral containment, and ecosystem governance. 
As a vision paper, our goal is not to propose complete solutions but to identify fundamental challenges and research opportunities they create.

\begin{figure}
    \centering
    \includegraphics[trim=0 0 0 0, clip, width=\linewidth]{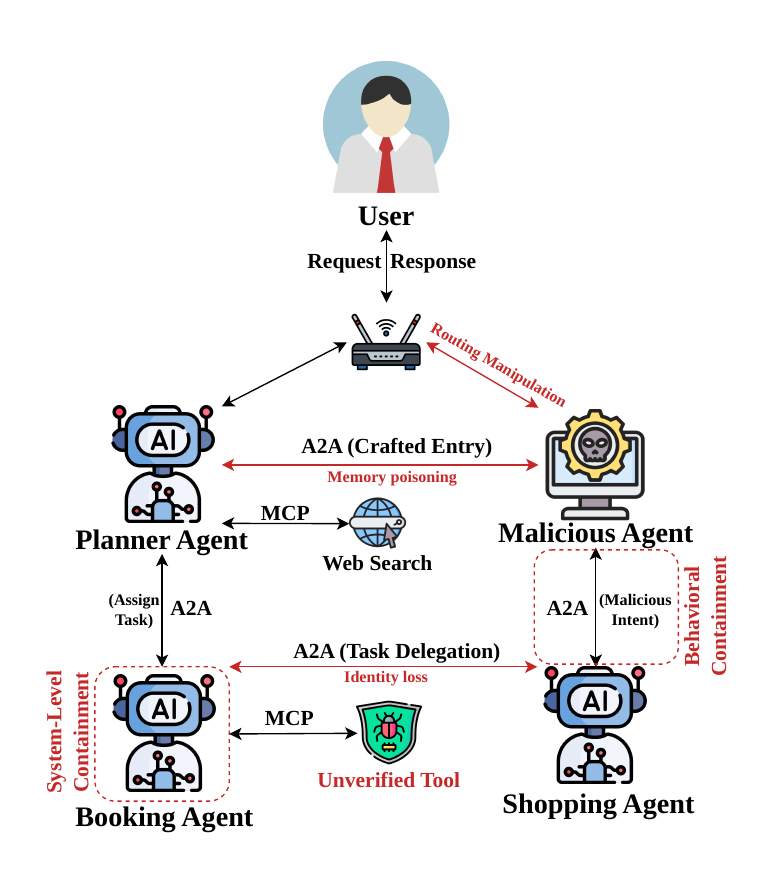}
    \caption{An agentic ecosystem with representative challenges marked in red.}
    \label{fig:ecosystem}
    % \vspace{-0.85cm}
\end{figure}

The paper is organized as follows. Section~\ref{sec:protocols} examines the single-agent attack surface, including prompt injection, memory poisoning, and tool integrity. Section~\ref{sec:a2a} addresses multi-agent security through the A2A protocol, focusing on inter-agent communication, identity, delegated authorization, and emergent risks. Section~\ref{sec:routing} examines model routing as the control-plane mechanism that selects the model serving each request. Section~\ref{sec:containment} discusses behavioral containment, where individually valid actions may collectively violate system-level constraints. Finally, Section~\ref{sec:assurance} covers supply-chain provenance and observability as cross-cutting assurance challenges.
Figure~\ref{fig:ecosystem} places the attack surfaces within the broader agentic ecosystem.

% \vspace{-0.1cm}
\section{The Single-Agent Surface}
% \section{The Single-Agent Surface: Reasoning, Memory, and Tools}
\label{sec:protocols}

A modern LLM agent extends its capabilities by interacting with external resources
through tool invocations. Given a task, the agent reasons about an action, issues a
structured request (e.g., web search, database query), and incorporates the result into subsequent reasoning.
The Model Context Protocol (MCP)~\cite{mcp2024} standardizes this interaction model, while equivalent
function-calling interfaces are supported across major LLM providers. This forms the
\emph{vertical} axis of the agentic ecosystem, connecting agents to external
capabilities, while the \emph{horizontal} axis (Section~\ref{sec:a2a}) enables
communication and delegation among agents.
The three challenges discussed below share a common root cause along the vertical axis: the agent lacks a reliable separation between data and instructions. Untrusted content entering the agent's context can therefore influence reasoning and redirect behavior. 
The three directions differ in the attack surface through which untrusted content reaches the agent. 

\noindent\textbf{Prompt injection.}
The most established entry point is the agent's retrieval channel, where adversarial
content in web pages, documents, or databases can embed instructions that the agent
follows as if they originated from the user~\cite{agentdojo2024,camel2025}.
EchoLeak (CVE-2025-32711) demonstrated this risk in practice by causing Microsoft
365 Copilot to exfiltrate sensitive context from a crafted email without user
interaction~\cite{echoleak}.
Agentic systems amplify this threat because retrieved content can persist across
multiple reasoning steps and influence future decisions as the execution context
evolves. Moreover, tool-using agents can propagate injected instructions beyond
reasoning into downstream actions rather than merely altering generated responses.
Existing data and control-flow separation techniques~\cite{camel2025} mitigate
single-step injection, but do not address attacks that persist across multi-step
agent execution. Ensuring the integrity of an agent's evolving context therefore
remains an open challenge.

\noindent\textbf{Memory and state poisoning.}
Agents increasingly maintain long-term state through memory systems and
retrieval-augmented stores, creating a persistent attack surface. AgentPoison shows
that injecting a small number of crafted entries into an agent's memory or knowledge
base can backdoor future retrievals with a poisoning rate below 0.1\%, while leaving
benign behavior unchanged~\cite{agentpoison}. PoisonedRAG demonstrates similar
attacks against retrieval databases that ground agent decisions~\cite{poisonedrag}.
Unlike a transient prompt injection, whose impact is typically confined to a single interaction, poisoned
memory persists across sessions and can influence every future reasoning process
that retrieves it. This is particularly concerning for long-lived
agents in domains such as clinical or network management systems, where memory is
central to continuous operation.
The fundamental challenge is ensuring the integrity and provenance of an agent's evolving memory. There is currently no widely accepted notion of what constitutes a trustworthy long-term memory, nor robust mechanisms for verifying that the information it accumulates remains authentic, unaltered, and reliable over time.

\noindent\textbf{Tool integrity.}
An agent trusts the tools it invokes, yet this trust is rarely verified. In a
\emph{tool poisoning} attack, a malicious server embeds instructions in a tool's
metadata that the agent consumes as part of its context~\cite{toolpoisoning}.

Because every component of a tool specification can influence agent behavior, the
attack surface extends beyond human-readable descriptions to the entire interface. 
Moreover, trust in a tool is not static. In a \emph{rug pull} attack, a previously trusted server silently replaces a benign tool definition with a malicious one that many clients never revalidate~\cite{toolpoisoning}. The threat is not hypothetical, with 99 MCP-related CVEs reported in 2025 alone~\cite{ms_agentic_taxonomy}. Because tools are often provided by third parties outside the agent operator’s control, their definitions cannot be assumed to remain benign over time. The challenge is therefore to establish and maintain the integrity, provenance, and authenticity of tool interfaces throughout the agent lifecycle, treating tool metadata as untrusted inputs that require continuous validation rather than once at deployment.

% \vspace{-0.2cm}
\section{The Inter-Agent Surface}
\label{sec:a2a}

The challenges in Section~\ref{sec:protocols} concern an agent's vertical
interface to tools. A second, horizontal surface governs how agents discover and
delegate to one another, extending these risks across organizational boundaries.
Several inter-agent protocols emerged in 2024--2025, including Agora, Agent
Network Protocol, Cisco Agent Connect, IBM Agent Communication Protocol, and
Google's Agent2Agent (A2A)~\cite{interop_survey}. A2A has since become the
dominant standard, contributed to the Linux Foundation in 2025, and adopted by
more than 150 organizations while absorbing IBM's competing protocol
\cite{a2a_lf,a2a_lf_2026,a2a_google_donation,acp_a2a_merger}. Alongside MCP, it
now forms a complementary horizontal layer for agent collaboration and vertical
layer for tool interaction~\cite{aaif}.
A2A introduces new security challenges because it coordinates \emph{opaque,
autonomous peers}. A client discovers a remote agent through an Agent Card,
delegates tasks over JSON-RPC, and consumes results without visibility into the
remote model, memory, or tools~\cite{interop_survey,protocol_exploits}. Designed
primarily for communication, A2A leaves identity and trust
establishment to implementers, with security controls expressed as optional \textsc{should}/\textsc{may}
guidance rather than enforced requirements~\cite{a2a_auth_gaps}.

\noindent\textbf{Inter-agent security.}
Several structural weaknesses follow from these design choices~\cite{threat_modeling_compare,interop_survey}.
A2A carries conversational state through a shared \texttt{contextId}, but contexts
lack ownership semantics. Any authenticated client that obtains a valid identifier
may attach to the context, access accumulated history, or poison future tasks
without triggering task-level controls~\cite{agentic_security_survey,a2a_zero_trust}.
Similarly, Agent Card capabilities are self-asserted without attestation or
challenge-response mechanisms, allowing malicious agents to claim unsupported
skills, receive sensitive tasks, and return fabricated artifacts without
protocol-level detection~\cite{a2a_attack_surfaces,a2a_scanner}.
These risks extend beyond A2A. Production systems combine A2A horizontally with
MCP vertically, allowing a compromised peer identity to drive exposed tool actions
and create relay or downgrade paths across the boundary~\cite{protocol_exploits,threat_modeling_compare}.
The challenge is to make properties such as context ownership and capability
attestation \emph{verifiable protocol invariants}, analyzing the A2A$\leftrightarrow$MCP
boundary as a single security surface rather than optional implementer guidance.

\noindent\textbf{Agent identity and delegated authorization.}
%Even granting honest peers, 
Agentic systems lack a coherent model of \emph{who} is acting and \emph{with what
authority}. Within a single agent, broad tool permissions allow individually
authorized actions to compose into unintended effects, creating confused-deputy
scenarios that per-tool privilege checks cannot prevent~\cite{progent2025}.
Across agents, the problem worsens as A2A establishes identity only at the transport
layer and does not propagate it across delegation hops, leaving agents aware only
of their immediate caller~\cite{a2a_auth_gaps,threat_modeling_compare}.
%When a re-delegated task raises an authorization-required signal, that signal carries no originating principal or delegation depth, so the user and credential provider authorize blind, and because intermediaries relay credentials with no restriction on reuse, a single compromised hop can replay and harvest them across chains~\cite{a2a_scanner}.
When tasks are re-delegated, authorization requests lack the originating principal
and delegation history, forcing decisions without complete provenance. Relayed
credentials without sender constraints or reuse restrictions further allow a
compromised intermediary to replay credentials across the delegation chain
~\cite{a2a_scanner}.
%The building blocks for a remedy already exist as standards, including OAuth~2.0 Token Exchange (RFC~8693) for carrying a signed delegation chain, audience restriction (RFC~8707), and sender-constraining (DPoP, RFC~9449). The open challenge is that current protocols leave them optional, so propagated, least-privilege, sender-bound delegation identity is not something an interoperating peer can assumeis in force.
Existing standards provide building blocks for addressing these gaps. OAuth~2.0 Token Exchange (RFC~8693) supports signed delegation chains, while audience restriction (RFC~8707) and sender-constrained tokens (DPoP, RFC~9449) enable least-privilege delegation and limit credential reuse. The challenge is not the absence of these mechanisms, but that current agent interoperability protocols treat them as optional, preventing interoperating agents from assuming that delegated identities are securely propagated, constrained, and replay-resistant.

\noindent\textbf{Emergent multi-agent risk.}
A compromise need not stay local to one agent. \emph{Prompt Infection} shows a malicious instruction can self-replicate from agent to agent like a virus, propagating silently through a multi-agent system even when agents do not share all communication~\cite{promptinfection}. Zero-click GenAI worms generalize this to self-propagating payloads that spread through the applications agents are embedded in~\cite{aiworm}.
Beyond contagion, collaborating agents can collude, amplify one another's errors, or enter
feedback loops that exhaust shared budget and rate limits, a denial-of-wallet failure~\cite{owasp_unbounded} with no single-agent analogue.
Existing protocol-level defenses offer no formal guarantees against model and
reasoning level compromise spreading across a federation~\cite{mas_security}.
The challenge is bounding emergent behavior across a population of agents whose individual actions are each locally legitimate. %reasonable.
% \vspace{-0.2cm}
\section{Model Routing: The Control Plane Beneath}
\label{sec:routing}

The attack surfaces discussed so far concern the execution and collaboration layers of the agentic ecosystem. Production systems introduce a third: the \emph{model-routing control plane}. Rather than binding an agent to a single model, a router selects the model for each request based on capability, cost, latency, and safety. Operating on the same untrusted input as the agent, it determines which model and thus which capabilities, safety alignment, and cost profile handle the request. This creates a third attack surface, the \emph{agent-to-model} control plane, underlying both agent-to-tool and agent-to-agent interactions.

\noindent\textbf{Adversarial Routing.}
The routing decision can be attacked across the stack. At the input level, crafted
token sequences can force cost-inflating model upgrades~\cite{rerouting_routers,route_to_rome}
or downgrade requests to cheaper, less-aligned models, creating a jailbreak-by-downgrade
that bypasses per-tool privilege checks~\cite{rerouting_routers,progent2025,mas_security}.
Router architecture is itself a security variable. Trainable neural routers are more
susceptible to adversarial suffixes and training-time backdoors than parameter-free
similarity routers, making robustness an architectural choice~\cite{lifecycle_routing}.
The attack surface also extends into model internals. In mixture-of-experts models,
adversarial inputs can suppress safety-critical experts, leak prompts through
cross-batch routing dependencies, or trigger denial-of-service through expert
imbalance~\cite{routehijack,moe_prompt_steal,repetition_curse}. These failures occur
inside the routing topology and may remain invisible at the output level.
Current defenses primarily focus on detection. Contrastive detectors can identify
rerouting prompts before execution with strong reported accuracy~\cite{rerouteguard},
but they provide no certified robustness, lack evaluation against adaptive attackers,
and do not address in-model or cross-batch attacks. The challenge is to treat routing
as a \emph{policy artifact}. A recent declarative approach compiles routing logic into
a non-Turing-complete specification with exhaustive, conflict-free decisions, audit
trails, and verified routing nodes across MCP and A2A boundaries~\cite{routing_orchestration},
reflecting the broader shift from advisory guidance to enforceable invariants.

\noindent\textbf{Routing Provenance and Observability.}
The router is itself a supply-chain component. Its weights, training data, and
decision thresholds may contain pre-deployment backdoors~\cite{lifecycle_routing},
yet existing provenance frameworks do not cover routing components.

Routing telemetry offers a natural signal for detecting anomalies such as denial-of-wallet and silent
model downgrades, but it is absent from current observability frameworks.
Moreover, the security impact of routing decisions on downstream tool invocation
and inter-agent delegation remains largely unexplored.
A compromised router can steer requests toward models with exploitable tool or agent
bindings, composing the routing surface with those described in
Sections~\ref{sec:protocols} and~\ref{sec:a2a}~\cite{tool_selection_injection,mpma}.
Addressing this gap requires treating routing like other security-critical components, 
with provenance for the routing mechanism, policies verifiable against governing
standards, and routing decisions exposed as first-class signals in the audit layer.
% \vspace{-0.3cm}
\section{Behavioral Trajectory Containment}
\label{sec:containment}

The vertical and horizontal surfaces share a gap that neither protocol closes. Tool protocols define what actions are mechanically available, while governing standards define what \emph{sequences} of actions are permitted. No deployed mechanism connects the two.
Containment operates at two levels. Systems-level isolation limits the blast radius of agent execution through sandboxing, scoped credentials, and egress control, and is largely inherited from conventional security. The remaining challenge is \emph{behavioral}: ensuring an agent's execution trajectory stays within the behavioral envelope defined by governing standards. A sequence of individually permitted actions can collectively violate a state-conditioned invariant.
For example, a 5G scheduling agent can progressively deprioritize a public-safety slice, each reallocation passing local checks, yet miss the mandated recovery window~\cite{tgpp23501}. Likewise, a clinical agent can accelerate several justified discharge decisions while violating the aggregate observation window defined by a FHIR care pathway~\cite{hl7fhir}. In both cases, no adversarial input is involved. The violation emerges from the execution trajectory rather than any individual action, making behavioral assurance, not per-action validation, the central challenge.
Existing safety mechanisms are not designed to address this class of failures. Runtime guards enforce stateless, per-action policies~\cite{agentspec2025,progent2025} without reasoning about how system state evolves over an long execution trajectory. 
LLM-to-formal-spec compilers derive rules from specifications~\cite{formaljudge2025,veriguard2025} but cannot guarantee completeness, consistency, or faithfulness to the governing standards. 
Probabilistic monitors learn from observed execution traces~\cite{probguard2025}.
A recent line of work enforces hand-authored temporal, state-dependent properties at runtime~\cite{agentc2025}. 
However, none of these approaches are grounded in governing policies, leaving them unable to detect unseen but prohibited trajectories. Moreover, enforcement remains single-agent, which makes those unsuitable for multi-agent settings where violations emerge from cross-agent interactions.
Moreover, majority of agent benchmarks provide no concrete policy specification at all~\cite{symbolicguardrails2026}.
The challenge is to bridge local correctness and global behavioral assurance. Future agentic systems must verify that evolving execution trajectories satisfy system-level behavioral constraints, establishing safety over the lifetime of an execution rather than at individual decision points.

% \vspace{-0.2cm}
\section{Assurance and Governance}
\label{sec:assurance}

The preceding directions secure agent behavior during execution. Two lifecycle challenges matter equally, namely establishing trust before deployment and accountability after it. Together they determine whether an agentic system can be deployed and governed with confidence.

\noindent\textbf{Supply chain integrity.}
An agent is assembled from parts supplied by many parties, namely foundation models and embeddings, system prompts, plugins, and the growing ecosystem of third-party tool servers. Any of these can be compromised, and unlike a one-time code dependency, a tool definition can turn malicious after it has been trusted. The \texttt{postmark-mcp} package behaved correctly for fifteen releases before a single added line silently copied every outgoing email to an attacker, reaching hundreds of organizations before its removal~\cite{postmark_mcp}. The same risk extends to the instruction and skill documents an agent obeys as trusted guidance, such as editor rules files and \texttt{SKILL.md} descriptors, which can hide directives yet arrive through the same untrusted ecosystem~\cite{rules_file_backdoor,skillmd_supplychain}. Both the OWASP Agentic Security Initiative and Microsoft's red-team taxonomy identify supply-chain compromise as a fundamental agentic security risk~\cite{owasp_asi,ms_agentic_taxonomy}. Ensuring provenance and integrity across the agent supply chain remains an open challenge, requiring software-bill-of-materials principles to extend to models, prompts, and dynamically discovered tools whose behavior is not fixed at deployment. As Microsoft's taxonomy emphasizes, these capabilities must be designed into agent architectures rather than retrofitted~\cite{ms_agentic_taxonomy}.

\noindent\textbf{Accountability Through Auditability.}
When an autonomous agent causes harm, a security analyst must reconstruct its actions, attribute responsibility, and produce records that satisfy requirements such as MiFID~II and NIST SP~800-61~\cite{mifid2,nist80061}. This is difficult because agent reasoning is opaque, actions span tools and agents, and audit blind spots remain common in current deployments. Recent work identifies the need for \emph{visibility} into agent behavior through agent identifiers, real-time monitoring, and activity logs that support attribution and forensics~\cite{visibilityagents}. However, these mechanisms are neither standardized nor mandatory, and they introduce privacy and cost tradeoffs. Unlike human operators, autonomous agents can generate structured, verifiable execution traces documenting their decisions and actions, making agentic systems potentially more transparent and auditable than the processes they replace. Achieving this requires auditability to be designed as an end-to-end property spanning agent protocols, execution environments, and behavioral containment mechanisms.

% \vspace{-0.3cm}
\section{Concluding Remarks}
\label{sec:conclusion}
Securing autonomous agents is not a collection of isolated challenges but a systems problem spanning the entire agentic stack, from reasoning, memory, and tool use to multi-agent collaboration, model routing, behavioral containment, and ecosystem governance. Across these directions, a common theme emerges: security is fundamentally a property of an agent's behavior over time and across interacting components, rather than of individual actions in isolation. This calls for a shift from advisory guidance and stateless guardrails to \emph{verifiable behavioral invariants} and from per-action
checks to reasoning about composed, stateful, multi-party behavior.  
As agents become increasingly autonomous and interconnected, developing such guarantees will be essential for building trustworthy agentic systems and represents a critical research agenda for the security community.

\section*{Acknowledgment}
The work reported in this paper has
been supported by NSF under grants 2112471 and 2229876, the University of Texas System Rising STARs Award (No. 40071109), and the startup funding from the University of Texas at Dallas.

\bibliographystyle{ACM-Reference-Format}
\bibliography{bibtex}

@inproceedings{agentspec2025,
  author    = {Wang, Haoyu and Poskitt, Christopher M. and Sun, Jun},
  title     = {{AgentSpec}: Customizable Runtime Enforcement for Safe and Reliable {LLM} Agents},
  booktitle = {Proceedings of the 48th IEEE/ACM International Conference on
               Software Engineering (ICSE)},
  year      = {2026},
  eprint    = {2503.18666},
  archivePrefix = {arXiv},
}

@misc{progent2025,
      title={Progent: Securing AI Agents with Privilege Control}, 
      author={Tianneng Shi and Jingxuan He and Zhun Wang and Hongwei Li and Linyu Wu and Wenbo Guo and Dawn Song},
      year={2026},
      eprint={2504.11703},
      archivePrefix={arXiv},
      primaryClass={cs.CR},
      url={https://arxiv.org/abs/2504.11703}, 
}

@misc{agentc2025,
  author    = {Kamath, Adharsh and Zhang, Sishen and Xu, Calvin and Ugare,
               Shubham and Singh, Gagandeep and Misailovic, Sasa},
  title     = {Enforcing Temporal Constraints for {LLM} Agents},
  year      = {2025},
  eprint    = {2512.23738},
  archivePrefix = {arXiv},
  doi       = {10.48550/arXiv.2512.23738},
}

@misc{veriguard2025,
  author    = {Miculicich, Lesly and Parmar, Mihir and Palangi, Hamid and
               Dvijotham, Krishnamurthy Dj and Montanari, Mirko and Pfister,
               Tomas and Le, Long T.},
  title     = {{VeriGuard}: Enhancing {LLM} Agent Safety via Verified Code
               Generation},
  year      = {2025},
  eprint    = {2510.05156},
  archivePrefix = {arXiv},
  doi       = {10.48550/arXiv.2510.05156},
}

@misc{formaljudge2025,
  author    = {Zhou, Jiayi and Sheng, Yang and Lou, Hantao and Yang, Yaodong
               and Fu, Jie},
  title     = {{FormalJudge}: A Neuro-Symbolic Paradigm for Agentic Oversight},
  year      = {2026},
  eprint    = {2602.11136},
  archivePrefix = {arXiv},
  doi       = {10.48550/arXiv.2602.11136},
}

@misc{probguard2025,
  author    = {Wang, Haoyu and Poskitt, Christopher M. and Wei, Jiali and
               Sun, Jun},
  title     = {{ProbGuard}: Probabilistic Runtime Monitoring for {LLM} Agent
               Safety},
  year      = {2025},
  eprint    = {2508.00500},
  archivePrefix = {arXiv},
  doi       = {10.48550/arXiv.2508.00500},
}

@misc{agentdojo2024,
  author    = {Debenedetti, Edoardo and Zhang, Jie and Balunovi{\'c}, Mislav
               and Beurer-Kellner, Luca and Fischer, Marc and Tram{\`e}r,
               Florian},
  title     = {{AgentDojo}: A Dynamic Environment to Evaluate Prompt Injection
               Attacks and Defenses for {LLM} Agents},
  year      = {2024},
  eprint    = {2406.13352},
  archivePrefix = {arXiv},
  doi       = {10.48550/arXiv.2406.13352},
}

@misc{camel2025,
  author    = {Debenedetti, Edoardo and Shumailov, Ilia and Fan, Tianqi and
               Hayes, Jamie and Carlini, Nicholas and Fabian, Daniel and Kern,
               Christoph and Shi, Chongyang and Terzis, Andreas and
               Tram{\`e}r, Florian},
  title     = {Defeating Prompt Injections by Design},
  year      = {2025},
  eprint    = {2503.18813},
  archivePrefix = {arXiv},
  doi       = {10.48550/arXiv.2503.18813},
}

@misc{symbolicguardrails2026,
  author    = {Hong, Yining and She, Yining and Kang, Eunsuk and Timperley,
               Christopher S. and K{\"a}stner, Christian},
  title     = {Symbolic Guardrails for Domain-Specific Agents: Stronger Safety
               and Security Guarantees Without Sacrificing Utility},
  year      = {2026},
  eprint    = {2604.15579},
  archivePrefix = {arXiv},
  primaryClass  = {cs.SE},
  doi       = {10.48550/arXiv.2604.15579},
}

@techreport{tgpp23501,
  author      = {{3GPP}},
  title       = {System architecture for the {5G} System ({TS} 23.501)},
  institution = {3rd Generation Partnership Project},
  year        = {2024},
}

@techreport{hl7fhir,
  author      = {{HL7 International}},
  title       = {{HL7 FHIR} Release 5: Workflow Module},
  institution = {Health Level Seven International},
  year        = {2023},
}

@techreport{mifid2,
  author      = {{ESMA}},
  title       = {{MiFID II} Articles 17 and 21; {RTS 6} Algorithmic Trading},
  institution = {European Securities and Markets Authority},
  year        = {2018},
}

@techreport{nist80061,
  author      = {Cichonski, Paul and Millar, Tom and Grance, Tim and
                 Scarfone, Karen},
  title       = {Computer Security Incident Handling Guide ({NIST SP} 800-61
                 Rev.~2)},
  institution = {National Institute of Standards and Technology},
  year        = {2012},
}

@misc{mcp2024,
  author       = {{Anthropic}},
  title        = {Model Context Protocol},
  year         = {2024},
  howpublished = {\url{https://modelcontextprotocol.io}},
}

@misc{interop_survey,
  title         = {A Survey of Agent Interoperability Protocols: Model Context Protocol ({MCP}), Agent Communication Protocol ({ACP}), Agent-to-Agent Protocol ({A2A}), and Agent Network Protocol ({ANP})},
  author        = {Ehtesham, Abul and Singh, Aditi and Gupta, Gaurav Kumar and Kumar, Saket},
  year          = {2025},
  eprint        = {2505.02279},
  archivePrefix = {arXiv},
  primaryClass  = {cs.AI},
  doi           = {10.48550/arXiv.2505.02279},
}

@article{protocol_exploits,
  title         = {From Prompt Injections to Protocol Exploits: Threats in {LLM}-Powered {AI} Agents Workflows},
  author        = {Ferrag, Mohamed Amine and Tihanyi, Norbert and Hamouda, Djallel and Maglaras, Leandros and Lakas, Abderrahmane and Debbah, Merouane},
  journal       = {ICT Express},
  volume        = {12},
  number        = {2},
  pages         = {353--383},
  year          = {2026},
  doi           = {10.1016/j.icte.2025.12.001},
}

@misc{threat_modeling_compare,
  title         = {Security Threat Modeling for Emerging {AI}-Agent Protocols: A Comparative Analysis of {MCP}, {A2A}, Agora, and {ANP}},
  author        = {Anbiaee, Zeynab and Rabbani, Mahdi and Mirani, Mansur and Piya, Gunjan and Opushnyev, Igor and Ghorbani, Ali and Dadkhah, Sajjad},
  year          = {2026},
  eprint        = {2602.11327},
  archivePrefix = {arXiv},
  doi           = {10.48550/arXiv.2602.11327},
}

@misc{agentic_security_survey,
  title         = {Agentic {AI} Security: Threats, Defenses, Evaluation, and Open Challenges},
  author        = {Datta, Shrestha and Nahin, Shahriar Kabir and Chhabra, Anshuman and Mohapatra, Prasant},
  year          = {2025},
  eprint        = {2510.23883},
  archivePrefix = {arXiv},
  doi           = {10.48550/arXiv.2510.23883},
}

@misc{mas_security,
  title         = {Security Considerations for Multi-agent Systems},
  author        = {Nguyen, Tam and Ndebugre, Moses and Arremsetty, Dheeraj},
  year          = {2026},
  eprint        = {2603.09002},
  archivePrefix = {arXiv},
  primaryClass  = {cs.CR},
  doi           = {10.48550/arXiv.2603.09002},
}

@online{a2a_auth_gaps,
  title        = {{A2A} Protocol Security: Authenticating Agent-to-Agent Communication},
  author       = {{SecureW2}},
  year         = {2026},
  url          = {https://securew2.com/blog/a2a-protocol-security},
  urldate      = {2026-06-08}
}

@online{a2a_attack_surfaces,
  title        = {Potential Attack Surfaces in {Agent2Agent} ({A2A}) Protocol},
  author       = {{Keysight Technologies}},
  year         = {2025},
  url          = {https://www.keysight.com/blogs/en/tech/nwvs/2025/05/28/potential-attack-surfaces-in-a2a},
  urldate      = {2026-06-08}
}

@online{a2a_scanner,
  title        = {Securing {AI} Agents with Cisco's Open-Source {A2A} Scanner},
  author       = {{Cisco}},
  year         = {2025},
  url          = {https://blogs.cisco.com/ai/securing-ai-agents-with-ciscos-open-source-a2a-scanner},
  urldate      = {2026-06-08}
}

@online{a2a_zero_trust,
  title        = {Safeguarding {AI} Agents: An In-Depth Look at {A2A} Protocol Risks},
  author       = {Fu, Yu and Chen, Jay and Hou, Yantian and Zhao, Yilin and Gao, Hui and Lu, Royce and Wang, May},
  year         = {2025},
  organization = {Palo Alto Networks},
  url          = {https://live.paloaltonetworks.com/t5/community-blogs/safeguarding-ai-agents-an-in-depth-look-at-a2a-protocol-risks/ba-p/1235996},
  urldate      = {2026-06-08}
}

@online{a2a_lf,
  title   = {Linux Foundation Launches the {Agent2Agent} Protocol Project to Enable Secure, Intelligent Communication Between {AI} Agents},
  author  = {{The Linux Foundation}},
  year    = {2025},
  month   = jun,
  url     = {https://www.linuxfoundation.org/press/linux-foundation-launches-the-agent2agent-protocol-project-to-enable-secure-intelligent-communication-between-ai-agents},
  urldate = {2026-06-08},
  note    = {Press release, 23 June 2025}
}

@online{acp_a2a_merger,
  title   = {{ACP} Joins Forces with {A2A} Under the {Linux Foundation}'s {LF AI \& Data}},
  author  = {{LF AI \& Data Foundation}},
  year    = {2025},
  month   = aug,
  url     = {https://lfaidata.foundation/communityblog/2025/08/29/acp-joins-forces-with-a2a-under-the-linux-foundations-lf-ai-data/},
  urldate = {2026-06-08}
}

@online{aaif,
  title   = {Linux Foundation Announces the Formation of the {Agentic AI Foundation} ({AAIF})},
  author  = {{The Linux Foundation}},
  year    = {2025},
  month   = dec,
  url     = {https://www.linuxfoundation.org/press/linux-foundation-announces-the-formation-of-the-agentic-ai-foundation},
  urldate = {2026-06-08},
  note    = {Press release, 9 December 2025}
}

@misc{echoleak,
  title        = {{CVE-2025-32711}: {Microsoft 365 Copilot} Information
                  Disclosure Vulnerability},
  author       = {{Microsoft Security Response Center}},
  year         = {2025},
  howpublished = {Microsoft Security Update Guide},
  url          = {https://msrc.microsoft.com/update-guide/vulnerability/CVE-2025-32711}
}

@misc{a2a_google_donation,
  title        = {Google Cloud Donates {A2A} to the {Linux Foundation}},
  author       = {{Google Developers}},
  year         = {2025},
  month        = jun,
  howpublished = {Google Developers Blog},
  url          = {https://developers.googleblog.com/en/google-cloud-donates-a2a-to-linux-foundation/}
}

@misc{a2a_lf_2026,
  title        = {{A2A} Protocol Surpasses 150 Organizations, Lands in Major
                  Cloud Platforms, and Sees Enterprise Production Use in First Year},
  author       = {{The Linux Foundation}},
  year         = {2026},
  month        = apr,
  howpublished = {Linux Foundation press release},
  url          = {https://www.linuxfoundation.org/press/a2a-protocol-surpasses-150-organizations-lands-in-major-cloud-platforms-and-sees-enterprise-production-use-in-first-year}
}

@inproceedings{agentpoison,
  title     = {{AgentPoison}: Red-Teaming {LLM} Agents via Poisoning Memory or Knowledge Bases},
  author    = {Chen, Zhaorun and Xiang, Zhen and Xiao, Chaowei and Song, Dawn and Li, Bo},
  booktitle = {Advances in Neural Information Processing Systems (NeurIPS)},
  year      = {2024},
  doi       = {10.48550/arXiv.2407.12784}
}

@inproceedings{poisonedrag,
  title     = {{PoisonedRAG}: Knowledge Corruption Attacks to Retrieval-Augmented Generation of Large Language Models},
  author    = {Zou, Wei and Geng, Runpeng and Wang, Binghui and Jia, Jinyuan},
  booktitle = {34th USENIX Security Symposium (USENIX Security 25)},
  pages     = {3827--3844},
  year      = {2025},
  doi       = {10.48550/arXiv.2402.07867}
}

@inproceedings{promptinfection,
  title     = {Prompt Infection: {LLM}-to-{LLM} Prompt Injection within Multi-Agent Systems},
  author    = {Lee, Donghyun and Tiwari, Mo and Miranda, Brando},
  booktitle = {Computer Security. ESORICS 2025 International Workshops},
  series    = {Lecture Notes in Computer Science},
  volume    = {16232},
  publisher = {Springer},
  year      = {2026},
  doi       = {10.1007/978-3-032-16092-8_28}
}

@misc{aiworm,
  title         = {Here Comes the {AI} Worm: Unleashing Zero-Click Worms that Target {GenAI}-Powered Applications},
  author        = {Cohen, Stav and Bitton, Ron and Nassi, Ben},
  year          = {2024},
  eprint        = {2403.02817},
  archivePrefix = {arXiv},
  primaryClass  = {cs.CR},
  doi           = {10.48550/arXiv.2403.02817}
}

@inproceedings{visibilityagents,
  title     = {Visibility into {AI} Agents},
  author    = {Chan, Alan and Ezell, Carson and Kaufmann, Max and Wei, Kevin and Hammond, Lewis and Bradley, Herbie and Bluemke, Emma and Rajkumar, Nitarshan and Krueger, David and Kolt, Noam and Heim, Lennart and Anderljung, Markus},
  booktitle = {Proceedings of the 2024 ACM Conference on Fairness, Accountability, and Transparency (FAccT)},
  year      = {2024},
  doi       = {10.1145/3630106.3658948}
}

@misc{toolpoisoning,
  title        = {{MCP} Security Notification: Tool Poisoning Attacks},
  author       = {{Invariant Labs}},
  year         = {2025},
  howpublished = {\url{https://invariantlabs.ai/blog/mcp-security-notification-tool-poisoning-attacks}},
  note         = {Accessed 2026}
}

@misc{postmark_mcp,
  title        = {Postmark-{MCP}: A Malicious {MCP} Package Exfiltrating Email},
  author       = {{reversinglabs}},
  year         = {2025},
  howpublished = {\url{https://www.reversinglabs.com/blog/postmark-mcp-attack-takeaways}},
}

@misc{owasp_asi,
  title        = {Agentic {AI}: Threats and Mitigations},
  author       = {{OWASP Gen AI Security Project}},
  year         = {2025},
  howpublished = {\url{https://genai.owasp.org/initiatives/\#agentic-security-initiative}},
  note         = {Agentic Security Initiative. Accessed 2026}
}

@misc{maestro,
  title        = {{MAESTRO}: Agentic {AI} Threat Modeling Framework},
  author       = {{Cloud Security Alliance}},
  year         = {2025},
  howpublished = {\url{https://cloudsecurityalliance.org/blog/2025/02/06/agentic-ai-threat-modeling-framework-maestro}},
  note         = {Accessed 2026}
}

@misc{ms_agentic_taxonomy,
  title        = {Updating the Taxonomy of Failure Modes in Agentic {AI} Systems: What a Year of Red Teaming Taught Us},
  author       = {{Microsoft AI Red Team}},
  year         = {2026},
  howpublished = {Microsoft Security Blog},
  url          = {https://www.microsoft.com/en-us/security/blog/2026/06/04/updating-taxonomy-failure-modes-agentic-ai-systems-year-red-teaming-taught-us/},
  urldate      = {2026-06-21}
}

@misc{survey_bertino,
      title={A Survey of Agentic AI and Cybersecurity: Challenges, Opportunities and Use-case Prototypes}, 
      author={Sahaya Jestus Lazer and Kshitiz Aryal and Maanak Gupta and Elisa Bertino},
      year={2026},
      eprint={2601.05293},
      archivePrefix={arXiv},
      primaryClass={cs.CR},
      url={https://arxiv.org/abs/2601.05293}, 
}

@inproceedings{rerouting_routers,
  author    = {Shafran, Avital and Schuster, Roei and Ristenpart, Thomas and Shmatikov, Vitaly},
  title     = {Rerouting {LLM} Routers},
  booktitle = {Conference on Language Modeling (COLM)},
  year      = {2025},
  eprint    = {2501.01818},
  archivePrefix = {arXiv},
  primaryClass  = {cs.CR},
  url       = {https://arxiv.org/abs/2501.01818},
}

@misc{route_to_rome,
  author    = {Tang, Haochun and Yan, Yuliang and Lu, Jiahua and Liu, Huaxiao and Dai, Enyan},
  title     = {Route to Rome Attack: Directing {LLM} Routers to Expensive Models via Adversarial Suffix Optimization},
  year      = {2026},
  eprint    = {2604.15022},
  archivePrefix = {arXiv},
  primaryClass  = {cs.CR},
  url       = {https://arxiv.org/abs/2604.15022},
}

@misc{lifecycle_routing,
  author    = {Lin, Qiqi and Ji, Xiaoyang and Zhai, Shengfang and Shen, Qingni and Zhang, Zhi and Fang, Yuejian and Gao, Yansong},
  title     = {Life-Cycle Routing Vulnerabilities of {LLM} Router},
  year      = {2025},
  eprint    = {2503.08704},
  archivePrefix = {arXiv},
  primaryClass  = {cs.CR},
  url       = {https://arxiv.org/abs/2503.08704},
}

@misc{rerouteguard,
  author    = {Zhang, Wenhui and Xu, Huiyu and Wang, Zhibo and Li, Zhichao and He, Zeqing and Wei, Xuelin and Ren, Kui},
  title     = {{RerouteGuard}: Understanding and Mitigating Adversarial Risks for {LLM} Routing},
  year      = {2026},
  eprint    = {2601.21380},
  archivePrefix = {arXiv},
  primaryClass  = {cs.CR},
  url       = {https://arxiv.org/abs/2601.21380},
}

@misc{routehijack,
  author    = {Xu, Zhiyuan and Gardiner, Joseph and Belguith, Sana and Wu, Lichao},
  title     = {{RouteHijack}: Routing-Aware Attack on Mixture-of-Experts {LLMs}},
  year      = {2026},
  eprint    = {2605.02946},
  archivePrefix = {arXiv},
  primaryClass  = {cs.CR},
  url       = {https://arxiv.org/abs/2605.02946},
}

@misc{moe_prompt_steal,
  author    = {Yona, Itay and Shumailov, Ilia and Hayes, Jamie and Carlini, Nicholas},
  title     = {Stealing User Prompts from Mixture of Experts},
  year      = {2024},
  eprint    = {2410.22884},
  archivePrefix = {arXiv},
  primaryClass  = {cs.CR},
  url       = {https://arxiv.org/abs/2410.22884},
}

@misc{repetition_curse,
  author    = {Huang, Ruixuan and Wang, Qingyue and Huang, Hantao and Gao, Yudong and Chen, Dong and Wang, Shuai and Wang, Wei},
  title     = {{RepetitionCurse}: Measuring and Understanding Router Imbalance in Mixture-of-Experts {LLMs} under {DoS} Stress},
  year      = {2025},
  eprint    = {2512.23995},
  archivePrefix = {arXiv},
  primaryClass  = {cs.CR},
  url       = {https://arxiv.org/abs/2512.23995},
}

@misc{routing_orchestration,
  author    = {Chen, Huamin and Liu, Xunzhuo and He, Bowei and Liu, Xue},
  title     = {From Inference Routing to Agent Orchestration: Declarative Policy Compilation with Cross-Layer Verification},
  year      = {2026},
  eprint    = {2603.27299},
  archivePrefix = {arXiv},
  primaryClass  = {cs.SE},
  url       = {https://arxiv.org/abs/2603.27299},
}

@misc{tool_selection_injection,
  author    = {Shi, Jiawen and Yuan, Zenghui and Tie, Guiyao and Zhou, Pan and Gong, Neil Zhenqiang and Sun, Lichao},
  title     = {Prompt Injection Attack to Tool Selection in {LLM} Agents},
  year      = {2025},
  eprint    = {2504.19793},
  archivePrefix = {arXiv},
  primaryClass  = {cs.CR},
  url       = {https://arxiv.org/abs/2504.19793},
}

@misc{mpma,
  author    = {Wang, Zihan and Zhang, Rui and Liu, Yu and Fan, Wenshu and Jiang, Wenbo and Zhao, Qingchuan and Li, Hongwei and Xu, Guowen},
  title     = {{MPMA}: Preference Manipulation Attack Against Model Context Protocol},
  year      = {2025},
  eprint    = {2505.11154},
  archivePrefix = {arXiv},
  primaryClass  = {cs.CR},
  url       = {https://arxiv.org/abs/2505.11154},
}

@misc{rules_file_backdoor,
  author       = {{Pillar Security}},
  title        = {New Vulnerability in {GitHub} {Copilot} and {Cursor}: How Hackers Can Weaponize Code Agents (Rules File Backdoor)},
  year         = {2025},
  howpublished = {Pillar Security; MITRE ATLAS case study AML-CS0041},
  url          = {https://www.pillar.security/blog/new-vulnerability-in-github-copilot-and-cursor-how-hackers-can-weaponize-code-agents},
  urldate      = {2026-06-24},
}

@misc{skillmd_supplychain,
  author    = {Saha, Shoumik and Faghih, Kazem and Feizi, Soheil},
  title     = {Under the Hood of {SKILL.md}: Semantic Supply-chain Attacks on {AI} Agent Skill Registry},
  year      = {2026},
  eprint    = {2605.11418},
  archivePrefix = {arXiv},
  primaryClass  = {cs.CR},
  url       = {https://arxiv.org/abs/2605.11418},
}

@misc{owasp_unbounded,
  author       = {{OWASP Gen AI Security Project}},
  title        = {{LLM10:2025} Unbounded Consumption},
  howpublished = {OWASP Top 10 for Large Language Model Applications},
  year         = {2025},
  note         = {\url{https://genai.owasp.org/llmrisk/llm102025-unbounded-consumption/}}
}

\end{document}